# Applied Explainability for Large Language Models: A Comparative Study

**Venkata Abhinandan** Kancharla
NRI Institute of Technology
abhinri6123@gmail.com
ORCID: https://orcid.org/0009-0000-2464-313X

**ABSTRACT**
Large language models (LLMs) achieve strong performance across many natural language processing tasks, yet their decision processes remain difficult to interpret. This lack of Transparency creates practical challenges when deploying models in real-world applications that require trust, debugging, and accountability. While numerous explainability methods have been proposed, their practical behaviour in transformer-based NLP systems is not always well understood.

This study presents an applied comparison of three explainability techniques—Integrated Gradients, attention-rollout, and SHAP on a fine-tuned DistilBERT model for SST-2 sentiment classification. Rather than proposing new methods, the focus is on evaluating how existing approaches behave under a consistent and reproducible setup. The methods were evaluated using qualitative criteria such as faithfulness, stability, and human interpretability.

The analysis shows that gradient-based attribution provides the most stable and intuitive explanations in this setting, while attention-based explanations are computationally efficient but less aligned with sentiment-relevant tokens. Model-agnostic explanations offer flexibility but introduce higher computational cost and variability. These findings highlight practical trade-offs that practitioners must consider when selecting explainability tools.

Overall, this work emphasises the importance of evaluating explainability methods in realistic usage scenarios and treating explanations as diagnostic aids rather than definitive accounts of model reasoning. The study provides practical insights for ML engineers working with transformer-based NLP systems and outlines directions for extending explainability analysis to larger models and diverse tasks.

## 1. Introduction

Large Language Models (LLMs) have transformed natural language processing by enabling systems to perform tasks such as sentiment analysis, question answering, and text generation with high accuracy. Transformer-based architectures, particularly BERT and its variants, learn rich contextual representations from large-scale data and have become central to modern NLP systems [3]. Despite their strong performance, these models operate through
complex internal mechanisms that are not easily understandable to humans. As a result, they are often described as opaque or black-box systems.

## 1.1 Motivation: Why LLMs are powerful but opaque

LLMs derive their power from large datasets, deep architectures, and attention-based mechanisms that capture subtle language patterns. However, the same complexity that enables high performance also makes it difficult to trace how predictions are formed. Prior work has shown that internal signals such as attention weights do not always correspond directly to human-understandable reasoning [10]. This creates a tension between model capability and transparency, especially when LLMs are deployed in decision support scenarios.

## 1.2 Why explainability matters in real-world ML systems

In applied machine learning, explainability is important for debugging, trust, and accountability. Practitioners often need to verify whether models rely on meaningful patterns rather than spurious correlations. Explainability tools help identify failure modes, detect bias, and support responsible deployment. As ML systems increasingly influence real-world decisions, the ability to inspect model behaviour becomes a practical requirement rather than a purely academic concern [14].

## 1.3 Gap between existing XAI methods and practical usage

A wide range of explainability methods has been proposed, including gradient-based attribution, attention visualisation, and model-agnostic approaches such as SHAP. While these methods are well studied, there remains a gap between theoretical proposals and their practical behaviour in transformer-based NLP systems. Some techniques may be computationally expensive, unstable across inputs, or difficult to interpret consistently. This gap highlights the need for applied evaluations that examine how explainability methods behave in realistic settings rather than idealised conditions.

## 1.4 Contributions of this paper

This work focuses on a practical comparison of explainability techniques applied to transformer-based language models. Instead of proposing a new method, the study evaluates how existing approaches behave in a controlled experimental setup.

The main contributions are:
- **A structured comparison of explainability methods for LLMs,** focusing on

attention-based, gradient-based, and feature attribution approaches.
- **An experimental evaluation on a downstream NLP** task using a fine-tuned transformer model.
- **Practical insights for ML/AI engineers** regarding the strengths and limitations of commonly used explainability tools.

## 2. Background and Preliminaries

Modern natural language processing systems are increasingly built on large-scale transformer-based models. As these models are adopted in real-world applications, concerns about transparency and decision-making have grown. This section provides background on large language models and key explainability concepts to situate this study.

### 2.1 Overview of Large Language Models

Large Language Models (LLMs) are neural networks trained on large text corpora to learn general language representations. Many LLMs are based on the **Transformer architecture**, which uses self-attention mechanisms to model relationships between tokens in a sequence [2].

Models such as **BERT** introduced bidirectional contextual learning, allowing token representations to depend on both left and right context [3]. Later, **DistilBERT** showed that smaller and more efficient models can retain much of the performance of larger architectures while reducing computational cost [4].

Because transformer-based models learn from large-scale data and complex parameter interactions, their internal decision processes are often difficult to interpret directly. This motivates the use of explainability techniques.

### 2.2 Explainability vs. Interpretability

The terms 'explainability' and 'interpretability' are related but not identical. Interpretability generally refers to how easily a human can understand the internal workings of a model. Explainability, in contrast, refers to methods that provide external insights into model behaviour without necessarily revealing internal mechanisms [12].

In many deep learning systems, models are not inherently interpretable due to their complexity. As a result, post-hoc explainability methods are often used to provide insight into predictions. Recognising this distinction helps clarify the role of explainability techniques in LLM research.

### 2.3 Types of Explainable AI

Explainable AI methods are often categorised based on how explanations are generated and what scope they cover.

**Post-hoc vs. Intrinsic:**
**Intrinsic interpretability** refers to models designed to be understandable by structure, such as linear models or decision trees. **Post-hoc explainability** methods, on the other hand, are applied after model training to analyse predictions without modifying the model [13]. Most explainability techniques used for LLMs belong to the post-hoc category.
**Local vs. Global Explanations:**
Local explanations focus on explaining individual predictions. For instance, token-level Attribution methods show which words influenced a specific output.
**Global explanations** describe overall model behaviour across many inputs, such as dataset-level patterns or model summaries [17].

In NLP explainability research, local explanations are commonly emphasised because individual Predictions are easier to inspect and validate.

## 3. Taxonomy of Explainability Techniques for LLMs

Explainability methods for large language models are commonly grouped based on how they derive explanations from model behaviour. Prior research in NLP explainability does not follow a single universal taxonomy, but several recurring categories appear in the literature. Broadly, These include attention-based methods, gradient-based attribution methods, feature attribution methods, and example-based explanations. Each group reflects a different way of interpreting how models process input text.

These categories are not strictly separated, and some methods overlap in ideas or combine multiple signals. However, this grouping provides a practical structure for understanding explainability techniques used with transformer-based language models.

### 3.1 Attention-based explanation methods

Attention-based methods rely on attention weights inside transformer models to visualise token interactions during prediction. Tools such as **BERTViz** allow researchers to inspect attention patterns across layers and heads, making it easier to observe contextual relationships between tokens[18]. Because attention values are directly accessible, these methods are easy to apply and computationally efficient.

However, the reliability of attention as an explanation has been questioned [10]. Showed that attention distributions do not always align with feature importance and can sometimes be altered without significantly affecting predictions. As a result, attention is often treated as a visualization aid rather than a definitive explanation, since prior work has shown that attention weights do not consistently correspond to feature importance in model predictions [10].

### 3.2 Gradient-based attribution methods

Gradient-based approaches estimate how sensitive predictions are to input tokens by analysing gradients with respect to embeddings. These methods attempt to measure how small input changes influence model outputs. **Integrated Gradients** is a widely used method in this category and attribute predictions relative to a baseline input [7].

Practical toolkits such as **Captum** make gradient-based attribution accessible for transformers models and are frequently used in applied research. Compared to attention-based methods, Gradient approaches are often considered more faithful because they directly track the model sensitivity to inputs.

### 3.3 Feature attribution methods

Feature attribution methods assign importance scores by modifying or masking parts of the input and observing prediction changes. These methods typically treat the model as a black box. **SHAP** is a well-known approach grounded in Shapley values from cooperative game theory [8].

SHAP has been applied to transformer models and is valued for theoretical grounding and model-agnostic design. At the same time, prior work notes that applying SHAP to long text Sequences can be computationally demanding, which limits large-scale use.

### 3.4 Example-based explanations

Example-based explanations justify predictions by identifying influential training instances. **TracIn** estimates the influence of individual training points by tracking their impact on the model parameters during training [11]. These methods are useful for dataset auditing, debugging, and understanding how models learn from data.

Although less common in routine NLP explainability compared to attribution methods, example-based approaches are increasingly relevant for data quality analysis and accountability.

### 3.5 Limitations of current taxonomies

Current taxonomies simplify a rapidly evolving field. Many explainability methods behave differently depending on model architecture, task type, and evaluation setup. Work on **Model Cards** emphasises documenting intended use and limitations when deploying models [15]. Evaluation studies in NLP have also shown that explanations can sometimes be unstable or misleading under certain conditions [20].

As LLMs grow in scale and usage, explainability research continues to evolve. Many studies recommend combining multiple approaches and interpreting explanations cautiously rather than relying on a single method.

## 4. Experimental Setup

This section describes the models, dataset, explainability methods, and evaluation criteria used in this study. All experiments were conducted using a fixed, fine-tuned model artifact to ensure reproducibility across explainability analyses. Code and trained models are publicly available at: https://github.com/abhinandan6123/Applied_XAI_for_LLMs

### 4.1 Model Architecture

We employ DistilBERT, a lightweight transformer-based language model derived from BERT, as the primary model for experimentation. **DistilBERT** retains much of BERT's representational capacity while significantly reducing computational overhead, making it well suited for applied and industry-orientated NLP tasks.

The model was fine-tuned for binary sentiment classification and subsequently frozen. The same trained model checkpoint was reused across all explainability experiments to ensure consistency.

### 4.2 Dataset

Experiments were conducted on the **SST-2 (Stanford Sentiment Treebank)** dataset, which consists of short English sentences annotated with binary sentiment labels (positive or negative). SST-2 is a widely adopted benchmark for evaluating sentiment classification models and is well suited for analysing token-level explanations due to its concise sentence structure. Only the classification task was considered, and no additional datasets were introduced during the explainability phase.

### 4.3 Explainability Methods

To analyse and compare different explainability paradigms, three complementary methods were Implemented:

#### 4.3.1 Integrated Gradients

Integrated Gradients was applied to compute token-level attribution scores by measuring the contribution of each input token to the model's prediction relative to a baseline. The resulting Attributions were visualised using bar plots, enabling direct inspection of sentiment-bearing tokens. Integrated Gradients served as the primary gradient-based explainability method in this study.

#### 4.3.2 Attention Rollout

Attention rollout was used to aggregate attention weights across transformer layers, producing a single attention distribution originating from the [CLS] token. This method provides a fast, attention-based explanation by propagating attention through residual connections. Attention Rollout was included to evaluate whether attention weights align with prediction-relevant Features.

### 4.3.3 SHAP
A Kernel SHAP–based approach was explored to obtain model-agnostic explanations. Due to practical constraints when applying SHAP to transformer-based text models, SHAP was configured using reshaped input representations and was used primarily for qualitative analysis.

SHAP outputs were generated and saved as HTML visualisations to support comparative analysis.

### 4.4 Evaluation Criteria
Explainability methods were evaluated qualitatively based on the following criteria:
- **Faithfulness**:
The extent to which highlighted tokens correspond to features that plausibly influence the model's prediction.
- **Stability**:
The consistency of explanations across repeated evaluations of similar or identical inputs.
- **Human Interpretability**:
The degree to which explanations align with human intuition and are easily understandable by practitioners.

These criteria were selected to emphasise practical usability and reliability over purely theoretical properties.

## 5. Results and Analysis
This section presents a comparative analysis of integrated gradients, attention rollout, and SHAP based on their observed behaviour during explainability experiments conducted on a fine-tuned DistilBERT model for SST-2 sentiment classification. The analysis focuses on stability, faithfulness, and practical usability, in line with the evaluation criteria defined in Section 4.

### 5.1 Quantitative Observations
While the primary objective of this work is not numerical benchmarking of explainability scores, Several consistent quantitative patterns were observed across runs:
- **Integrated Gradients (IG)** produced relatively stable token-level attribution scores across repeated evaluations of similar inputs. Attribution magnitudes showed limited variance when the same sentence or semantically similar sentences were analysed multiple times.

- **SHAP**, implemented using a Kernel SHAP–based approach, exhibited noticeable variability across runs. Small changes in input representation and background data configuration led to changes in attribution distributions, indicating sensitivity to input perturbations and configuration choices.
- **Attention Rollout** was computationally efficient and executed significantly faster than both IG and SHAP. However, the resulting attention distributions did not consistently correlate with the model's prediction-relevant features, limiting their quantitative reliability as faithful explanations.

Overall, IG demonstrated the most consistent quantitative behaviour among the three methods, while SHAP showed higher variance and attention-rollout prioritised efficiency over faithfulness.

### 5.2 Qualitative Analysis

Qualitative inspection of saved attribution visualisations revealed clear differences in how each method explains model behaviour:

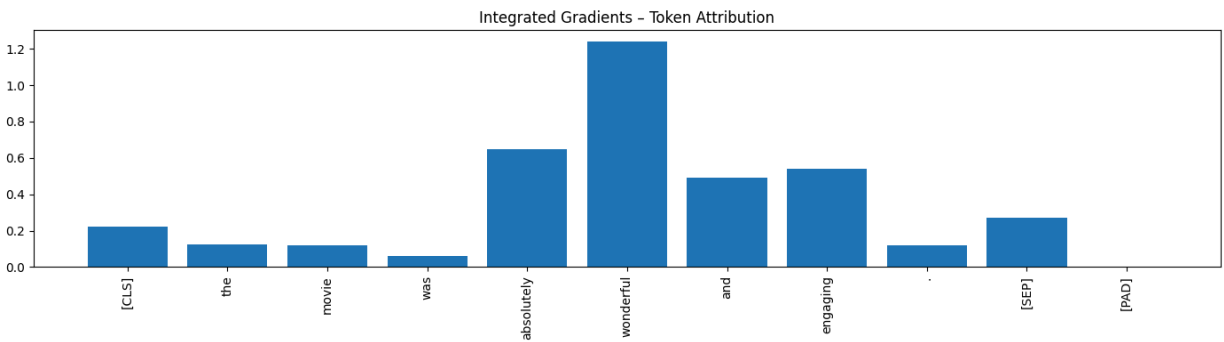

**Figure 1:** Integrated Gradients attribution for a positive review. Sentiment-bearing words such as *"wonderful"* and *"engaging"* receive higher importance scores.

- **Integrated Gradients** consistently highlighted sentiment-bearing tokens such as adjectives, negations, and intensifiers (e.g., words contributing directly to positive or negative sentiment). These attributions aligned well with human intuition and remained consistent across multiple examples.

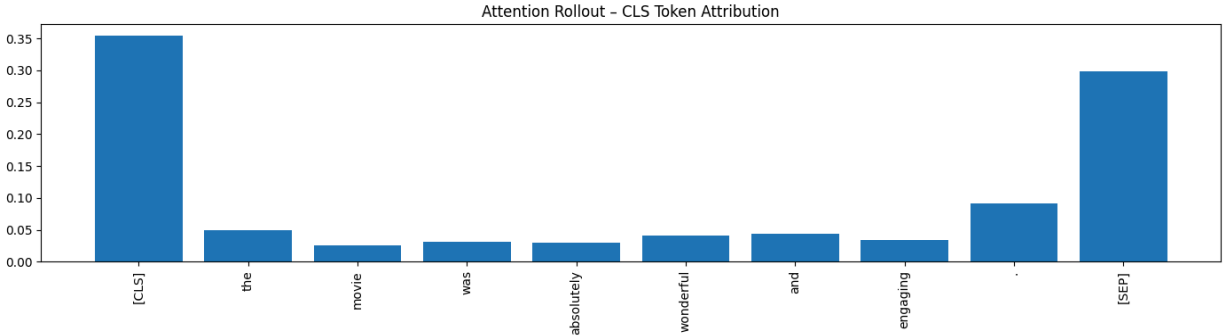

**Figure 2:** Attention rollout attribution showing higher focus on the [CLS] token and structural tokens

● **Attention Rollout** frequently emphasised syntactic or structural tokens, including stopwords, punctuation, and positional tokens (e.g., the [CLS] token). In several cases, sentiment-relevant words received comparatively lower attention weights, reducing interpretability from a human perspective.

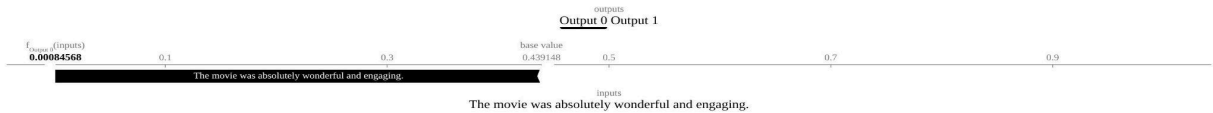

**Figure 3:** SHAP token attribution for a sentiment classification example.

● **SHAP explanations,** when successfully computed, identified sentiment-relevant input components but often appeared noisy at the token level. The explanations were less visually stable than IG outputs and required careful preprocessing and configuration to interpret meaningfully.

These observations suggest that IG provides clearer and more intuitive explanations for sentiment classification compared to Attention Rollout and SHAP.

**5.3 Failure Cases**

Each explainability method exhibited specific failure modes during experimentation:
- **Attention-based explanations** occasionally emphasised irrelevant or weakly relevant tokens, even when the model prediction was correct. This behaviour reinforces prior findings that attention weights do not always correspond to causal or faithful explanations.
- **SHAP explanations** were sensitive to small changes in input formatting and background data selection. In some cases, this sensitivity resulted in unstable attribution patterns, making large-scale or repeated analysis difficult without extensive tuning.
- **Integrated Gradients**, while generally stable, relies on gradient information and thus remains dependent on model smoothness and baseline selection. However, no severe Failure cases were observed during the conducted experiments.

These failure cases highlight the importance of understanding methodological limitations when applying explainability tools in practice.

### 5.4 Trade-off Summary
Based on the experimental results and observations, the following trade-offs were identified:

**Table 1: Comparison of Explainability Methods**

| Methods | Strengths | Limitations | Practical Usefulness |
|---|---|---|---|
| **Integrated Gradients** | High faithfulness and stable explanations; moderate computational cost | Requires gradient access and baseline selection | Well suited for practical debugging and analysis of transformer-based NLP models |
| **SHAP** | Flexible and model-agnostic framework | Higher computational overhead and instability in transformer-based text settings | Limited scalability and reliability in practice |
| **Attention Rollout** | Fast and easy to compute | Often less aligned with prediction-relevant features | Less reliable as a standalone interpretability method |

Overall, Integrated Gradients emerged as the most practically reliable technique in this study, while SHAP and Attention Rollout provided complementary perspectives, revealed notable limitations.

## 6. Discussion
### 6.1 Implications for Industry ML Systems
The experimental results highlight important considerations for deploying explainability techniques in real-world machine learning systems. In practical ML workflows, explainability is often required for debugging, model validation, and building stakeholder trust rather than for academic analysis alone. In this context, Integrated Gradients demonstrated the most reliable behaviour by consistently highlighting tokens that aligned with the model's predictions and human intuition. This reliability makes gradient-based methods more suitable for production environments where engineers need explanations that are both stable and interpretable.

Attention Rollout, while computationally efficient, showed limitations when applied as a standalone explanation technique. Its tendency to focus on structural tokens reduces its usefulness for debugging or decision justification in industry settings. SHAP, although flexible and model agnostic, introduced significant computational and implementation overhead, which can be challenging in time-sensitive or resource-constrained systems. These observations suggest that Explainability methods must be evaluated not only by their theoretical appeal but also by their operational practicality.

### 6.2 When Explainability Helps and When It Fails
Explainability proved most useful when attribution methods aligned closely with the underlying decision-making process of the model. Integrated Gradients helped surface meaningful token-level contributions, enabling a clearer understanding of why certain predictions were made. This is particularly valuable when diagnosing misclassifications or verifying that a model is relying on reasonable features.

However, the experiments also showed that explainability can be misleading when methods are applied without careful interpretation. Attention-based explanations frequently emphasised tokens that were not semantically relevant to sentiment, which could lead to incorrect assumptions about model behaviour. Similarly, SHAP explanations exhibited sensitivity to input representation and sampling choices, which reduced their stability across similar inputs. These Failure cases reinforce the need for cautious interpretation and highlight that explainability Outputs should not be treated as definitive explanations without contextual validation.

### 6.3 Recommendations for Practitioners
Based on the experimental findings, practitioners should prioritise explainability methods that balance faithfulness, stability, and ease of use. Gradient-based approaches such as Integrated Gradients are well suited for transformer-based NLP models when consistent and interpretable

Explanations are required. Attention-based methods may still be useful for exploratory analysis but should not be relied upon as sole indicators of model reasoning. Model-agnostic techniques like SHAP can provide complementary insights but are better suited for targeted qualitative analysis rather than routine deployment.

Overall, practitioners should view explainability as a diagnostic tool rather than a definitive explanation of model behaviour. Combining multiple methods, validating explanations against domain knowledge, and understanding method-specific limitations are essential steps toward responsible and effective use of explainability in applied machine learning systems.

## 7. Limitations and Threats to Validity

Like any applied study, this work has limitations that should be considered when interpreting the results. These limitations do not invalidate the findings but clarify the scope within which they should be understood.

### 7.1 Dataset Limitations

This study uses the SST-2 sentiment classification dataset, which contains short, single-sentence movie reviews labeled with binary sentiment. While SST-2 is a standard benchmark in NLP, it represents a narrow domain and relatively simple text structure compared too many real-world language tasks.

Because of this, the observed behaviour of explainability methods may not fully generalise to longer documents, multi-label tasks, or specialised domains such as medical or legal text. The dataset also focuses on English sentiment classification, which limits conclusions about multilingual or domain-specific scenarios.

### 7.2 Model Scope Limitations

The experiments are conducted on DistilBERT, a compact transformer model designed for efficiency. Although DistilBERT retains strong performance, it is smaller than many modern large language models.

As a result, explanation behaviours observed here may differ when applied to larger or more recent models. This study does not evaluate extremely large-scale models or instruction-tuned LLMs, and therefore conclusions should be interpreted within the context of medium-scale transformer models.

### 7.3 Human Interpretation Bias

Explainability methods ultimately rely on human interpretation of outputs such as highlighted tokens or attribution scores. Different users may interpret the same explanation differently depending on their expectations or domain knowledge.

Additionally, explanations that appear intuitive to humans are not always guaranteed to reflect true model reasoning. This study focuses on practical observation rather than formal human evaluation studies, which introduces subjectivity in assessing explanation quality.

Recognising this limitation is important when using explainability methods for decision support.

## 8. Conclusion and Future Work

This study presented an applied comparison of explainability techniques for transformer-based language models, with a focus on practical usability rather than theoretical novelty. Using a fine-tuned DistilBERT model on a standard sentiment classification task, we evaluated Integrated Gradients, Attention Rollout, and SHAP under a consistent experimental setup.

### 8.1 Summary of findings

The experimental results showed clear differences in how these methods behave in the real-world settings. Integrated Gradients consistently produced stable and intuitive token-level explanations that aligned well with sentiment-bearing words, making it the most reliable technique among those evaluated. Attention-rollout was computationally efficient and easy to apply but frequently emphasised structural or syntactic tokens, limiting its usefulness as a faithful explanation method. SHAP, while flexible and model-agnostic, introduced significant practical challenges in transformer-based NLP tasks, including higher computational cost and sensitivity to input representation, which constrained its applicability to qualitative analysis.

### 8.2 Lessons for applied ML engineers

From an applied perspective, these findings highlight that not all explainability methods are equally suitable for large language models in production or research workflows. Methods that are theoretically appealing may still face practical limitations when applied to modern transformer architectures. As demonstrated in this study, gradient-based approaches such as Integrated Gradients offer a favourable balance between faithfulness, stability, and usability for NLP practitioners.

### 8.3 Future directions

Future work can extend this analysis to larger models, additional downstream tasks, and more diverse datasets to assess the scalability of explainability techniques. Exploring explainability in multimodal models and developing tools that better align attention mechanisms with causal Explanations also remain important directions. Overall, this work emphasises the importance of evaluating explainability methods not only by their theoretical foundations but by their practical behaviour in real-world machine learning systems.